\documentclass[10pt,twocolumn,letterpaper]{article}

\usepackage{cvpr}
\usepackage{times}
\usepackage{epsfig}
\usepackage{graphicx}
\usepackage{amsmath}
\usepackage{amssymb}
\usepackage{subfiles}
\usepackage{booktabs}
\usepackage{float}

\usepackage[pagebackref=true,breaklinks=true,letterpaper=true,colorlinks,bookmarks=false]{hyperref}

\cvprfinalcopy % *** Uncomment this line for the final submission

\def\cvprPaperID{****} % *** Enter the CVPR Paper ID here

\ifcvprfinal\fi
\begin{document}

%%%%%%%%% TITLE
\title{Recognizing Manipulation Actions from State-Transformations}

\author{Nachwa Aboubakr\\
    \small Univ. Grenoble Alpes, Grenoble INP,\\
    \small CNRS, Inria, LIG \\
    {\tt\small nachwa.aboubakr@inria.fr}\\
\and
James L. Crowley\\
    \small Univ. Grenoble Alpes, Grenoble INP,\\
    \small CNRS, Inria, LIG \\
    {\tt\small james.crowley@inria.fr}\\
\and
Remi Ronfard\\
    \small Univ. Grenoble Alpes, Inria, \\
    \small Grenoble INP, CNRS, LJK \\
    {\tt\small remi.ronfard@inria.fr}
}
\maketitle
%\thispagestyle{empty}

%-------------------------------------------------------------------------
\begin{abstract}
Manipulation actions transform objects from an initial state into a final state.
In this paper, we report on the use of object state transitions as a mean for recognizing manipulation actions. 
Our method is inspired by the intuition that object states are visually more apparent than actions from a still frame and thus provide information that is complementary to spatio-temporal action recognition.
We start by defining a state transition matrix that maps action labels into a pre-state and a post-state.
From each keyframe, we learn appearance models of objects and their states.
Manipulation actions can then be recognized from the state transition matrix.
We report results on the EPIC kitchen action recognition challenge.
  %In this paper, we study the task of recognizing manipulation actions from a small number of frames. 
  %Manipulation actions transform object states from an initial state to a final state. 
  %Our method is based on the intuition that we can learn object states more efficiently than actions from a single frame.
  %We can better describe object states from still images than actions. 
  %We start by defining a state transition matrix that maps action labels into a pre-state and a post-state. Then, 
  %we adapt our previous model that learns object states to recognize actions from the transformation of object states. 
  %we then learn appearance models of object states in each frame, then recognize actions from the state transition matrix.
  %We report results on the EPIC kitchen action recognition challenge. 
  %In this paper, we study the task of recognition of manipulation actions as state transformation. 
  %We adapt our previous model that learns object states to recognize actions from the transformation of object states. 
  %we formulate the problem as a problem of recognizing transformation in the state of objects in a scene.

\end{abstract}
%-------------------------------------------------------------------------
\section{Introduction}
%Along with the success of object recognition task in computer vision, the field of action recognition has gained recently enormous attention. 

Most current approaches to action recognition interpret a frame sequence as a spatio-temporal signal.
%include an adoption of successful object recognition systems. 
%interpret a spatio-temporal signal. 
3D Convolutional Neural Networks are a direct adaptation of 2D CNN to the spatio-temporal case.
However, it %\cite{Carreira2017} 
results in a substantial increase in the number of parameters that must be learned, greatly increasing the computational cost and the requirements for training data. 
An alternative approach is to exploit object recognition models and follow 2D spatial kernels by either 1D temporal kernels (2.5D ConvNets) \cite{xie2018rethinking}, or a Recurrent Neural module \cite{donahue2015long}.
Researchers have also explored the use of two-stream networks in which one stream is used to analyze image appearance from RGB images and the other represents motion from optical flow maps \cite{wang2016temporal, simonyan2014two, kalogeiton2017joint}. 
These approaches provide spatio-temporal analysis while avoiding the very large increase in learned parameters.
%??
%The approach described below can provide a complementary source of information for such analysis. 

%- learn object's state transition throughout the video and not video motion.
An alternative to learning spatio-temporal models for action recognition from video is to learn relations between entities from a sequence of frames \cite{Ma2017, Baradel_2018_ECCV}.
Baradel et al. \cite{Baradel_2018_ECCV} proposed a convolutional model that is trained to predict both object classes and action classes in two branches. This model is followed by an object relation network that learns to reason over object interactions. 
%%??
%These related works learn activities and objects from a sequence of frames. 

\begin{figure}[t]
\centering
   \includegraphics[width=\linewidth]{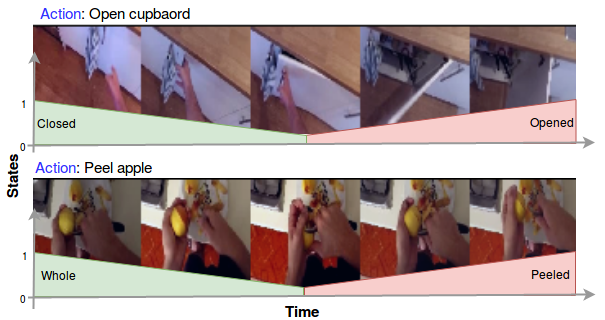}
   \caption{Changes in object states over time for action recognition. Two sample sequences from the EPIC kitchen dataset.} 
    \label{fig:frames_sample}
\end{figure}
\begin{figure*}[ht]
\centering
    \includegraphics[width=\linewidth]{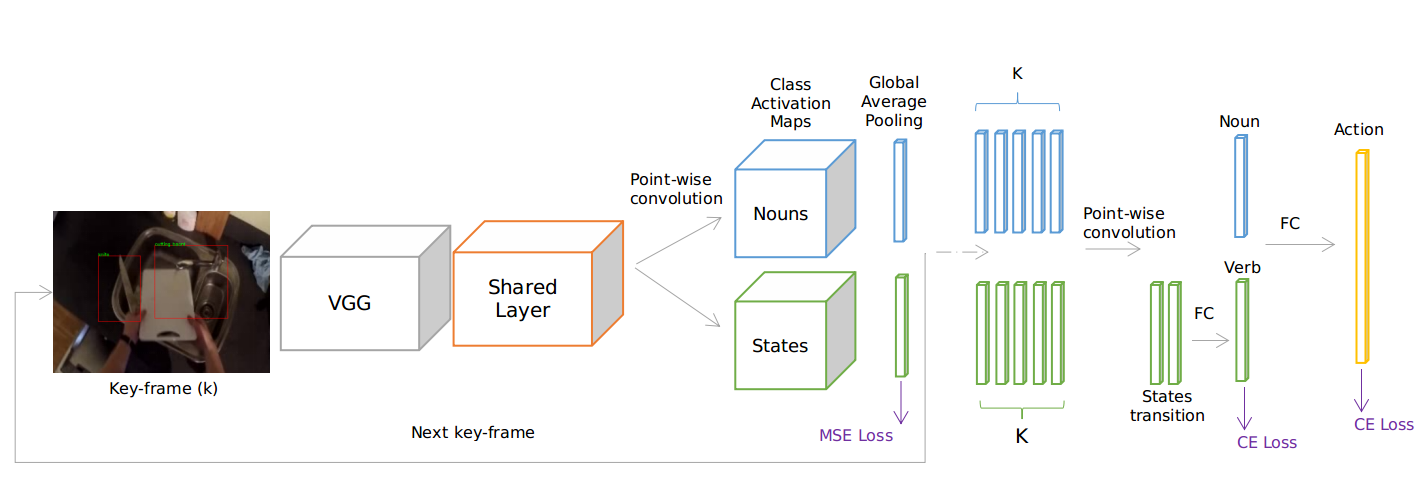}
    \caption{Proposed architecture of learning action recognition as state transformations.}
    \label{fig:noun_state_arch}
\end{figure*}
These analyses are inspired from human ability 
%Humans are able
to develop an understanding of a situation using a limited number of static observations. Human associate these observations with background knowledge in a form of previously seen episodes or past experience \cite{fleuret2011comparing, blusseau2014psychophysical}.
This ability allows a human subject to interpret a complex scene from static images and make hypotheses about unseen actions that may have occurred and could explain changes to the scene.
For example, we can understand which action is shown in Figure ~\ref{fig:frames_sample} with 5 keyframes or less from the video clip. 
Inferring the associated actions in frame sequences is a relatively effortless task for a human, while it remains challenging for machines \cite{Stabinger2016}.
%-action recognition
We believe that such analysis may provide an effective method for inferring actions from a set of frames which are chronologically ordered and contains semantic relations between objects. Such inference would complement hypotheses from spatio-temporal action recognition.

% how this different than us, 
A manipulation action transforms an object from a pre-existing state (pre-state) into a new state (post-state). Thus we can say that the action \textit{causes} a change in the state of the corresponding object. %% object or entity - be consistent
Alayrac et al. \cite{alayrac2017joint} have investigated automatic discovery of both object states and actions  from videos. They treat this problem as a discriminative clustering problem by exploiting the ordering of the frames. Their work is promising, even though it has been studied only on a small number of action classes. %(7 actions). 

In this paper, we propose to train a model to recognize object types and object states from a small number of frames and then use changes in object states to predict actions. Our intuition is that object types and states are visually more apparent from a single frame than an action verb. 
%Thus we explore the task of action recognition as a state transformation of objects that appear in a sequence of keyframes. 

%-------------------------------------------------------------------------
\section{Manipulation action as state transformation}

%- define action & action attributes
An action, as defined in the Cambridge dictionary\footnote{Cambridge University Press. (2019). Cambridge online dictionary, Cambridge Dictionary online. Retrieved at April 3, 2019}, is the effect something has on another thing.
Therefore, a manipulation action $a_i \in A$ is composed of:
the subject that performs the action, 
the verb $v_i \in V$ which describes the effect of the action, 
and the object $n_i \in N$ the effect is applied to.
%The hypothesis is that a human manipulation action can be represented as sequence of state changes to objects under manipulation.

%- set of nouns and a verb, New datasets has been released to address this task. (cite 50salads and MPII kitchens as one block task).  
The action recognition problem can be formulated with one class for each possible combination of these attributes. For example, \textit{cut tomato} and \textit{cut cucumber} can be considered as two different classes as in \cite{stein2013combining}. 
Some recent datasets have considered the decomposition of an action into a verb and one or more objects $a = (v, (n_1, .., n_n))$ \cite{gao2019vrkitchen, Damen2018EPICKITCHENS, li2018eye}.
This makes it possible to study the task of action recognition as a composition of several sub-tasks (e.g. object detection and action verb recognition). 
%-------------------------------------------------------------------------
\subsection{State-changing actions}

%- our modelling of an action as (objects and states)
%Manipulation actions, or actions for short, can be defined as the set of actions that are performed while interacting with objects with the intention of performing a change of their state. 
We are concerned with recognizing manipulation actions that change the state of objects $s_i \in S$. 
%We study the task of recognizing these actions by recognizing the transition of object's states. 
The state change can appear in the object's shape, its appearance (color), or its location. Examples of object states include: closed, opened, full, empty, whole, and cut. 

%- Mapping between object states (pre and post) state. Each verb transforms an object from one state to another.
We define a state transition function $F$ that transforms the corresponding object from a pre-state $s_i$ into a post-state $s_j$. 
In some cases, this state transition can be defined directly from the type of action verb $v_i$. 
On the other hand, we have noticed that sometimes a single verb is not enough to distinguish an action.
For example, the verb \textit{remove} can mean open in \textit{remove lid} and can mean peel in \textit{remove the skin of the garlic}. Therefore, the state transition must take into account both action verbs and nouns.

%- state value depends on frames position in video.
Since the state changes happen as we move through time, the transition function $F$ returns a real value of each state depending on the frame position in the video segment. As in Figure~\ref{fig:frames_sample} the object starts in its initial state that gradually fades out and the post-state starts to appear as we advance in the video. 
%The same logic applies to the post-state which starts to appear in the later parts of the video and at the end the object becomes completely in its post-state. 
We suppose that the state changing frame is the mid-frame of the video action segment. 
Therefore, we define the action transition mapping function $F(v, n)$, which takes the action's verb $v$ and a set of objects (nouns) $n$ and returns a continuous value of objects' states for each frame depending on the frame position in the video. 
For example, the action \textit{open fridge} changes the fridge state from opened to closed. 

\begin{table*}[ht]
\centering
    \begin{tabular}{@{}rcccc|cccc@{}}
    %\toprule
              & \multicolumn{4}{c|}{Seen kitchens subset (S1)}                         
              & \multicolumn{4}{c}{Unseen kitchens subset (S2)}      \\ 
              & Acc T1    & Acc T5    & Precision      & Recall         
              & Acc T1    & Acc T5    & Precision      & Recall \\ \cmidrule(l){2-9} 
              & \multicolumn{8}{c}{Action}                  \\ \cmidrule(l){2-9} 
    Our model(RGB)     
                & 19.76  & 36.98  & 9.83 & \textbf{10.23}
                & 9.08  & 19.46 & 3.68  & 4.77  \\
2SCNN\cite{simonyan2014two}(RGB)    
                & 13.67  & 33.25  & 6.66  & 5.47  
                & 6.79  & 20.42 & 3.39  & 3.01  \\
%2SCNN\cite{simonyan2014two}(Fusion) 
%                & 13.23  & 30.36  & 5.35  & 4.46 
%                & 7.31  & 19.49 & 2.86  & 2.69  \\
TSN\cite{wang2016temporal}(RGB)      
                & \textbf{19.86}  & \textbf{41.89}  & \textbf{9.96}  & 8.81  
                & \textbf{10.11} & \textbf{25.33} & \textbf{4.77}  & \textbf{5.67}  \\
%TSN\cite{wang2016temporal}(Fusion)   
%                & \textbf{20.54}  & 39.79  & \textbf{11.57} & 9.78  
%                & \textbf{10.89}  & 25.26  & \textbf{6.21}  & \textbf{6.49}   \\ 
\cmidrule(l){1-9} 

              & \multicolumn{8}{c}{Verb}                        \\ \cmidrule(l){2-9} 
Our model(RGB)        
                & \textbf{47.41}  & 81.33  & 31.20 & 20.43 
                & 34.35  & 69.24 & 15.09 & 11.00 \\
2SCNN\cite{simonyan2014two}(RGB)    
                & 40.44  & 83.04  & 33.74 & 15.9  
                & 33.12 & 73.23 & 16.06 & 9.44  \\
%2SCNN\cite{simonyan2014two}(Fusion) 
%                & 42.16  & 80.58  & 29.39 & 14.83 
%                & 36.16 & 71.97 & 18.11 & 10.52 \\
TSN\cite{wang2016temporal}(RGB)      
                & 45.68  & \textbf{85.56}  & \textbf{61.64} & \textbf{23.81} 
                & \textbf{34.89} & \textbf{74.56} & \textbf{19.48} & \textbf{11.22} \\
%TSN\cite{wang2016temporal}(Fusion)   
%                & \textbf{48.23}  & 84.09  & 47.26 & 22.33 
%                & \textbf{39.4}  & 74.29 & \textbf{22.54} & \textbf{13.06} \\
\cmidrule(l){1-9} 
              & \multicolumn{8}{c}{Noun}                    \\ \cmidrule(l){2-9} 
Our model(RGB)      
                & 28.31 & 53.77  & 21.21 & 22.48 
                & 17.48 & 37.56  & 10.71 &12.55 \\
2SCNN\cite{simonyan2014two}(RGB)    
                & 30.46  & 57.05  & 28.23 & 23.23 
                & 17.58 & 40.46 & 11.97 & 12.53 \\
%2SCNN\cite{simonyan2014two}(Fusion) 
%                & 29.14  & 53.7   & 30.73 & 21.1  
%                & 18.03 & 38.41 & 15.31 & 12.55 \\
TSN\cite{wang2016temporal}(RGB)      
                & \textbf{36.8}   & \textbf{64.19}  & \textbf{34.32} & \textbf{31.62} 
                & \textbf{21.82} & \textbf{45.34} & \textbf{14.67} & \textbf{17.24} \\
%TSN\cite{wang2016temporal}(Fusion)   
%                & 36.71  & 62.32  & \textbf{35.42} & 30.53 
%                & \textbf{22.7}  & \textbf{45.72} & \textbf{15.33} & \textbf{17.52} \\ 
\cmidrule(l){1-9} 

    %\bottomrule
    \end{tabular}
    \caption{Results on the EPIC kitchen dataset (Seen and Unseen subsets). Highest values are in bold. Results of baseline methods (2SCNN and TSN) are reported by \cite{Damen2018EPICKITCHENS}.}
    \label{tab:res_epic}
\end{table*}

%-------------------------------------------------------------------------
\subsection{Architecture}

%- so our task is to identify a noun and a state for each key-frame. 
%Our objective is to recognize the manipulation actions by recognizing the transition of affected object states. 
In previous work \cite{Aboubakr2018}, we investigated detection and location of object types as well as object states from images. In this paper, we extend this work to learn changes in object state from keyframes.
%- overview of the architecture
%We propose to start by learning objects' (type and state) separately from each key-frame. After that, we perform later fusion of these concepts over key-frames.
The architecture of our model is shown in Figure~\ref{fig:noun_state_arch}. Given a video segment, we first split it into $k$ sub-segments of equal length and sample a random keyframe from each sub-segment. 
%This results in one clip of a sequence of $k$ keyframes.
For each keyframe, we learn two conceptual classes (object types and object states) separately. 
Then, from the selected sequence of $k$ keyframes, we extract two channels using a point-wise convolution from which we construct the state transition matrix (pre-state, post-state). For object types (nouns), we use a point-wise convolution to extract a vector of nouns that appear in the video segment. 
Action verbs are then learned from the state transition matrix. 
In the end, the action classes are learned directly from the set of object types and action verbs.
% Check to see if what I have said is accurate !!!
%- Architecture in details (show CAMs images) We expect that the model learns two concepts from each key-frame (object types and object states) 

% Un comment here for foveal vision part
%\subfile{foveal_vision}
%-------------------------------------------------------------------------
\section{Implementation details}

\paragraph{EPIC Kitchen dataset.} 
We study state transformations through the action labels in the EPIC Kitchen dataset. The EPIC Kitchen dataset is a large dataset of egocentric videos of people cooking and cleaning.
In this dataset, an action label is composed of a tuple of $a_i = $ (verb $v_i$, noun $n_i$) extracted from a narrated text given for each video action segment.

The EPIC verb represents the action verb while the EPIC noun is the action object. 
As the EPIC Kitchen dataset is an egocentric dataset which suggests one subject in the scene, 
the action subject is always the cook's hands.
We group each action verb depending on the type of effect they cause into 3 different groups: those that change the object's shape, color appearance, or location. This study leaves some non-state-changing verbs (like the verb \textit{check}) out of those groups as it does not change any object states. 
%The dataset has 125 different action verbs and 
As a result we define 49 state transitions and 31 different states.
\vspace{-10pt}
\paragraph{Network Architecture.}
As shown in Figure~\ref{fig:noun_state_arch}, we use a similar setting as in \cite{Aboubakr2018} for each keyframe. We start by extracting deep features using a VGG16 network with batch normalization \cite{simonyan2014very} pre-trained on ImageNet dataset \cite{deng2009imagenet}. 
In building our network architecture, we considered to use few trainable parameters. Thus, VGG layers are frozen during the whole training process. VGG features provide the input to a shared\footnote{shared over the two attributes (object types and states)} $3\times3$ convolutional layer. 
We separate the learning of object attributes into two branches: one for object types and the other for object states. Each attribute is learned with an independent loss. 

%- late fusion
For each keyframe, one noun vector and one state vector are extracted using Global Average Pooling over corresponding Class Activation Maps.
Afterwards, we perform a point-wise convolution to extract one noun vector and the states transition matrix over keyframes. 
Verbs are learned directly from the state transition matrix using a fully-connected (FC) layer. 
Both action attributes (verb, nouns) are fused using at a late stage a FC layer for action classification. 
All hidden layers use the ReLU (rectified linear unit) activation function.
A frame can have one or more states and/or nouns. Therefore, we treat nouns and states as multi-label classification problems that are learned with a Mean Square Error (MSE). On the other hand, verbs and actions are learned with a Cross Entropy (CE) function.
\vspace{-10pt}
\paragraph{Training.}
We use EPIC Kitchen video segments for training our model\footnote{Code is available at \url{https://github.com/Nachwa/object_states}}.
%Each video segment is split into equally spaced sub-segments. From each sub-segment, we choose a random keyframe for training. 
A clip is a collection of $k$ randomly sampled keyframes from $k$ equal length sub-segements, and it represents the corresponding action video segment. This strategy has been used in multiple works with similar problems \cite{wang2016temporal, Baradel_2018_ECCV}.
\vspace{-10pt}
\paragraph{EPIC challenge evaluation.} 
For evaluation, we aggregate the results of 10 clips as in \cite{Baradel_2018_ECCV}. We report the same evaluation metrics provided by the EPIC challenge \cite{Damen2018EPICKITCHENS}. Provided metrics include class-agnostic and class-aware metrics; Top-1 and Top-5 micro-accuracy in addition to precision and recall over only many shot classes (i.e. classes with more that 100 samples).

%-------------------------------------------------------------------------
\section{Discussion}
We report the primary results of the model in Table~\ref{tab:res_epic} from frames on the EPIC Kitchen dataset for action recognition task. As the test sets are not publicly available yet, we compared our results to two baseline techniques, 2SCNN model \cite{simonyan2014two} and TSN model \cite{wang2016temporal}, as reported in \cite{Damen2018EPICKITCHENS}. 
%Table~\ref{tab:res_epic} shows results from one modality (RGB) as well as from fusion of RGB and Flow inputs.

In our model, we only use RGB channels. The model has 20M parameters and only 5M trainable parameters which is significantly lower than both baseline techniques i.e. for each input modality: 2SCNN model \cite{simonyan2014two} uses 170M parameters and TSN model \cite{wang2016temporal} has 11M trainable parameters.
%On the other hand, our model is trained from a small number of frames. 
Even though, our model outperforms 2SCNN model \cite{simonyan2014two} in most of reported metrics while results of verbs and actions recognition are still comparable to TSN reported results\cite{wang2016temporal}.
%In a future work, we plan to improve our model embedding of objects (nouns). 
%as well as evaluate state transitions on more datasets. 
%We believe that the main contribution of this work is the learn of object states from frames instead of action verbs directly.
\vspace{-4pt}
%-------------------------------------------------------------------------
\section{Conclusion}

In this paper, we investigated a method for recognition of manipulation actions. 
The method proposes the recognition of changes of objects attributes from a small set of keyframes.
%rather than from spatio-temporal patterns 
We demonstrate that this can provide efficient recognition of manipulation actions.
%as transitions between successive frames. 
%Our method is based on the fact that an object and its state are the more apparent/describable attributes from a single frame than an action. 
%Therefore, we showed a first attempt to tackle this problem using a simple model that learns objects and their states independently and uses state transition to infer about action. 
We propose a model of manipulation action recognition from state changes that is conceptually sound and efficient.
We reported results of our model on the challenge of EPIC kitchen dataset and compare these to two baseline techniques. 
For the action recognition task, our model outperforms one of the baseline techniques using $34$ times less training parameters, and achieved comparable results with another of the baselines.
%Our model is learned on 2 times less training parameters 

%In the future, we plan to improve exploit the idea to improve our model and perform a more extensive comparison.
\vspace{-7pt}
%-------------------------------------------------------------------------

{\small
\bibliographystyle{ieee_fullname}
\bibliography{main}
}

\end{document}